\documentclass[11pt, twoside]{article}
\usepackage{fullpage,ifthen,epsf,hyperref,url,breakurl,enumerate,comment,flushend,multirow}
\usepackage[active]{srcltx}
\usepackage{amsthm,amsfonts,amsmath,amssymb,mathtools}
\usepackage{microtype,multicol,sidecap,mathrsfs}
\usepackage{esint}
\usepackage{algorithm}
\usepackage[noend]{algpseudocode}
\usepackage[usenames,dvipsnames,svgnames,table]{xcolor}
\usepackage{color}

\usepackage{graphicx}

\usepackage{tabularx}
\usepackage{epsfig}
\usepackage[english]{babel}
\usepackage[utf8]{inputenc}
\usepackage{dsfont}
\usepackage{pdfpages}
\usepackage{array}
\usepackage{eqparbox}
\usepackage{subfigure}
\usepackage{siunitx}

\newlength{\widebarargwidth}
\newlength{\widebarargheight}
\newlength{\widebarargdepth}

\makeatletter
\long\def\@makecaption#1#2{
        \vskip 0.8ex
        \setbox\@tempboxa\hbox{\small {\bf #1:} #2}
        \parindent 1.5em 
        \dimen0=\hsize
        \advance\dimen0 by -3em
        \ifdim \wd\@tempboxa >\dimen0
                \hbox to \hsize{
                        \parindent 0em
                        \hfil 
                        \parbox{\dimen0}{\def\baselinestretch{0.96}\small
                                {\bf #1.} #2
                                } 
                        \hfil}
        \else \hbox to \hsize{\hfil \box\@tempboxa \hfil}
        \fi
        }
\makeatother

\newtheorem{theorem}{Theorem}[section]

\newtheorem{definition}[theorem]{Definition}

\title{Symmetry Learning for Function Approximation in Reinforcement Learning}
\author{
Anuj Mahajan$^{\dagger}$ and Theja Tulabandhula$^{\ddagger}$\\
$^{\dagger}$Conduent Labs India ; $^{\ddagger}$University of Illinois Chicago\\
\texttt{$^{\dagger}$anujmahajan.iitd@gmail.com ; $^{\ddagger}$tt@theja.org }
}
\date{February 27, 2017}
\begin{document}

\maketitle
\begin{abstract}
In this paper we explore methods to exploit symmetries for ensuring sample efficiency in reinforcement learning (RL), this problem deserves ever increasing attention with the recent advances in the use of deep networks for complex RL tasks which require large amount of training data. We introduce a novel method to detect symmetries using reward trails observed during episodic experience and prove its completeness. We also provide a framework to incorporate the discovered symmetries for functional approximation. Finally we show that the use of potential based reward shaping is especially effective for our symmetry exploitation mechanism. Experiments on various classical problems show that our method improves the learning performance significantly by utilizing symmetry information.
\end{abstract}

\section{Introduction}
\begin{figure}
\centering
\includegraphics[width=0.7\textwidth]{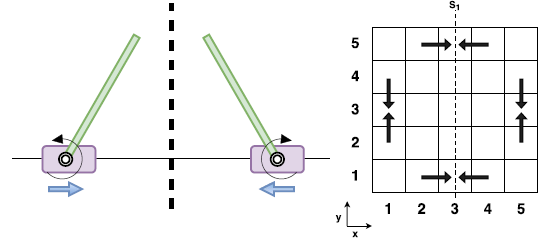}
\caption{\emph{Left}: Symmetry in Cart-Pole environment. \emph{Right}: Symmetry in grid world. $f((x,y))=(6-x,y)$, $g_s:\{N\rightarrow N,S\rightarrow S,E\rightarrow W,W\rightarrow E\}\forall s$.}
\label{cps}
\end{figure}

Reinforcement Learning (RL) is the task of training an agent to perform optimally in an environment using the reward and observation signals perceived upon taking actions which change the environment dynamics. Learning optimal behavior is inherently difficult because of challenges like credit assignment and exploration-exploitation trade offs that need to be made while converging to a solution. In many scenarios, like training a rover to move on a Martian surface, the cost of obtaining samples for learning can be high (in terms of robot's energy expenditure etc.), and so sample efficiency is an important subproblem which deserves special attention. 
Very often it is the case that the environment has intrinsic symmetries which can be leveraged by the agent to improve performance and learn more efficiently. For example, in the Cart-Pole domain ~\cite{barto1983neuronlike,sutton1998book} the state action space is symmetric with respect to reflection about the plane perpendicular to the direction of motion of the cart (Figure~\ref{cps}). In fact, in many environments, the number of symmetry relations tend to increase with the dimensionality of the state space. For instance, for the simple case of grid world of dimension $d$ (Figure ~\ref{cps}) there exist $O(d!2^d)$ fold symmetries. This can provide substantial gains in sample efficiency while learning as we would ideally need to consider only the equivalence classes formed under the induced symmetry relations.
However, discovering these symmetries can be a challenging problem owing to noise in observations and complicated dynamics of the environment. \par With recent advances in deep reinforcement learning~\cite{mnih2015human,levine2013guided}, it has been demonstrated that a lot of seemingly complex tasks like game play for Atari, control in simulated physics environments etc., which pose challenges in the form of large (possibly continuous) state action spaces and the difficulty of learning good representations~\cite{bengio2013representation}; can be handled very well with the use of deep networks as functional approximators. Training these networks, however requires large amounts of data and learning their parameters  necessitates coming up with careful update procedures and defining the right objectives(see ~\cite{glorot2010understanding}) and is a topic of study in its own right. This points us to the fact that it is essential to come up with methods that ensure sample efficiency in function approximation based RL, which builds up the premise for this work.

 To the best of our knowledge, we are the first to motivate the use of symmetry in the aforementioned context. In this paper we investigate methods for discovering state space symmetries and their inclusion as prior information via a suitable cost objective. We also show that our method dovetails seamlessly with the framework of using potential based reward shaping ~\cite{ng1999policy} which can help reduce the size of the data structures required for symmetry detection and provides additional information for establishing robust similarity estimates. Experiments on various classical control settings validate our approach with agents showing significant gains in sample efficiency and performance when using symmetry information.

\section{Related Work }
Model minimization ~\cite{zinkevich2001symmetry,dean1997model} in MDP literature is a closely related field in which symmetries are defined using equivalence relations on state and state action spaces, but a more generalized notion of symmetries is presented in ~\cite{ravindran2001symmetries} which allows for greater state space reduction. However, these approaches require that symmetries in environment be explicitly stated and do not handle discovery. Methods like ~\cite{girgin2010improving,Girgin} use conditionally terminating sequences formed from observed reward action sequences and directly try to estimate state equivalence and are closely related to our methods for symmetry detection, however they fail to be of use even in relatively simple environments like Cart-Pole where the action mappings under the symmetry transformation are not invariant. Most of these methods have prohibitively high overheads, moreover none of them address aforementioned issues in the context of using them for function approximation in deep RL. In many realistic scenarios the tasks needed to be learned by the RL agent is composed of several subtasks which form a hierarchy across the state space. Methods that try to use this intrinsic structure fall in the framework of temporally abstract actions~\cite{mcgovern1998macro,sutton1999between}. Although, these methods require extensive domain knowledge specification by the agent's designer.~\cite{stolle2002learning,mcgovern2002autonomous} have tried to automate this process by learning from agent observations but their approach is suboptimal as they fail to discover all the abstractions~\cite{Girgin}. Recent approaches for incorporating TAA use linear function approximation ~\cite{szepesvari2014universal,sorg2010linear} and have been shown to be effective. Value function generalization for sub goal setting by \cite{schaul2015universal} also gives better generalization over unseen sub goals in the function approximation setting. Our method has the potential to be augmented with these works for finding symmetries in sub-goal spaces. 

\section{Preliminaries}
In this section we give preliminary definitions and provide an overview of the topics we will build upon. \\

\noindent\textbf{Symmetries in MDP}: We represent MDP as a tuple $M:=\langle S,A,\Psi,T,R \rangle$  where $S$ is the set of states, $A$ is the set of actions, $\Psi \subset S\times A$ is the set of admissible state-action pairs, $R : \Psi\times S \rightarrow \mathbb{R}$, $T : \Psi\times S \rightarrow [0, 1]$ are reward and transition functions respectively. The notion of symmetries in MDP can be rigorously treated using the concept of MDP \textit{homomorphisms}(see Appendix ~\ref{app:defs} for some definitions). \cite{ravindran2001symmetries} define an MDP homomorphism $h$ from $M=\langle S,A,\Psi,T,R \rangle$ to $M'=\langle S',A'\allowbreak,\Psi',T',R' \rangle$  as a surjection $h:\Psi\rightarrow \Psi'$, which is itself defined by a tuple of surjections $\langle f,\{g_s,s\in S\}\rangle$. In particular, $h((s,a)):=(f(s),g_s(a))$, with $f:S\rightarrow S'$ and $g_s:A_s\rightarrow A_{f(s)}'$, which satisfies two requirements: Firstly it preserves the reward function (i.e., $R'(f(s),g_s(a),f(s'))=R(s,a,s')$) and secondly it commutes with transition dynamics of $M$ (i.e., $T'(f(s),\allowbreak g_s(a),f(s'))=T(s,a,[s']_{B_{h|S}})$). Here we use the notation $[\cdot ]_{B_{h|S}}$ to denote the \textit{projection} of equivalence classes $B$ that partition $\Psi$ under the relation $h((s,a))=(s',a')$ on to $S$. 
Symmetries $\chi:\Psi \rightarrow \Psi$ can then be formally defined as \textit{automorphisms} on $M$ that completely preserve the system dynamics with the underlying functions $f,g_s$ being bijective. The homomorphism requirements for a symmetry reduce to:
\begin{eqnarray}
\label{symtr}
T(f(s),g_s(a),f(s'))=T(s,a,s'), \textrm{ and}\\ 
\label{symrr}
R(f(s),g_s(a),f(s'))=R(s,a,s').
\end{eqnarray} 
The set of \textit{equivalence classes} $C[(s,a)]$ of state action pairs formed under the relation $\chi((s,a))=(s',a')$ (or more generally under any homomorphism $h$) partition $\Psi$ and can thus be used to form a quotient MDP, represented as $M_Q=M/C$, which is smaller and can be efficiently solved. As an example in Figure~\ref{cps}, for symmetry $\chi_{S1}$ w.r.t vertical axis, we have equivalence class of state action pair $[((2,1),E)]_{\chi_{S1}}=\{((2,1),E),((4,1),W)\}$
However in the RL setting, we do not know the underlying system dynamics and consequently, we do not know $C$ in advance, thus we cannot perform a model reduction. A workaround for this would be to estimate $C[(s,a)]$ on the go using the agent's experience and use the estimated set of equivalent classes to drive identical updates for equivalent state-action pairs during the process of learning the optimal policy.\\

\noindent\textbf{Reward Shaping}:
Reward shaping is a technique that augments the reward function $R$ of a MDP $M=\langle S,A,\Psi,T,R \rangle$ by a shaping function $F:\Psi\times S\rightarrow \mathbb{R}$, with the goal of providing the agent with additional useful information about the environment in the form of rewards. Thus, the agent now sees a modified reward $R'(s,a,s') = R(s,a,s') + F(s,a,s')$ when it takes actions. The notion of reward shaping in RL has its roots in behavioral psychology~\cite{skinner1990behavior}. It has been shown that the shaping is helpful in learning optimal policies if the shaping rewards are carefully designed to ensure that the policy invariance property holds~\cite{randlov1998learning}.~\cite{ng1999policy} have shown that if the reward shaping function is potential based i.e., is of the form: $F(s,a,s') = \gamma\Theta(s')-\Theta(s) \;\;\forall s,a,s'$ for some $\Theta:S\rightarrow \mathbb{R}$, then the policy invariance property is guaranteed. As we shall see in next section not only does shaping help in faster convergence to optimal policy, it can be crucial for detecting symmetries in the environment as-well. 

\section{Methodology}
In this section we present our approach for simultaneously discovering and exploiting symmetries in the RL framework. Our goal is to incorporate the state action space symmetry information early on during the learning so that \emph{regret} can be minimized and the optimal policy can be learned in a sample efficient manner. The main components involved in our method are: (a) symmetry detection, and (b) learning with symmetry based priors. Although our work can be applied to any form of functional approximation, we focus on $Q$ function approximation using deep feed forward networks in our study to illustrate the merits of our solutions. We elaborate each of these components next.\\

\noindent\textbf{Symmetry Detection}: Given an MDP $M=\langle S,A,\Psi,T,R \rangle$, we define set $\Pi_{sa,j}=\{(\sigma,N_{\sigma})\}$ where $\sigma$ is a sequence of rewards of length $j$ and $N_{\sigma}$ is the number of times it is seen starting with state $s$ taking action $a$  during the execution of policy $\pi$. We use the notation $|\Pi_{sa,j}| = \sum_{|\sigma|=j}N_\sigma$ and $\Pi_{sa,j}\cap \Pi_{s'a',j} = \{(\sigma,\min(N_\sigma,N'_\sigma))\}$. We define the notion of similarity between two state action pairs $\langle s,a \rangle$  and $\langle s'a'\rangle$ as follows:
\small
\begin{align}
\label{similarity}
\chi_{i,l_0}(\langle s,a \rangle ,\langle s'a'\rangle)=\frac{\sum_{j=l_0}^i|\Pi_{sa,j}\cap \Pi_{s'a',j}|}{(\sum_{j=l_0}^i|\Pi_{sa,j}|*\sum_{j=l_0}^i|\Pi_{s'a',j}|)^{1/2}}.
\end{align}
\normalsize
To efficiently compute the similarities between all the state action pairs, we use an auxiliary
structure called the reward history tree $P$, which stores the prefixes of reward
sequences of length up to $i$ for the state action pairs observed during policy execution. 
A reward history tree $P(N,E)$ is a labeled rooted tree (root being a null node). Here, $N$ is the set of nodes with each node labeled with a reward observation. And $E$ is the set of directed edges defined in the following way: Let the sequence of reward labels obtained while traversing to $n$ starting from the root be denoted by $\sigma_n$, then directed edge $(n,n')\in E$ if and only if $\sigma_n$ appended with the reward label of $n'$ form a prefix for some observed reward sequence. Additionally each node $n$ maintains a list of state, action, occurrence frequency tuples $[\langle s,a,o \rangle]$, Thus node $n$,  will store a tuple $\langle\hat{s},\hat{a},\hat{o}\rangle$ if the reward sequence $\sigma_n$ was observed for $\hat{o}$ times starting from $\hat{s}$ taking action $\hat{a}$  during policy execution. The similarities can then be computed by doing a breadth first traversal on  $P$ and maintaining two arrays $A_u(\langle s,a \rangle)$ and $ A_p(\langle s,a \rangle ,\langle s'a'\rangle)$ that store the occurrences and co-occurrences of the observed reward sequences. An estimate can be computed by: $\chi_{i,l_0}(\langle s,a \rangle ,\langle s'a'\rangle)=\frac{A_p(\langle s,a \rangle ,\langle s'a'\rangle)}{(A_u(\langle s,a \rangle) \cdot A_u(\langle s',a' \rangle))^{1/2}}$. 

We consider state pairs \begin{displaymath}
\chi_{sym} :=\{\langle s,a \rangle ,\langle s'a'\rangle|\chi_{i,l_0}(\langle s,a \rangle ,\langle s'a'\rangle)\geq \Delta\},
\end{displaymath}  as similar for the given length ($i$ and $l_0$) and threshold parameters ($\Delta$).
We thus propose that the symmetry of an environment can be deduced from the reward structure. This assumption is backed by the fact that many real life applications do provide rich information in the reward dynamics and are amenable to our method for discovering symmetries. A very simple example would be a robot trying to approach a well defined object placed at a distance, the fraction of pixels covered by the object on robot's visual sensors would provide an approximate notion of distance of robot from the object and is thus a good reward signal. Moreover it will allow to discern which locations are equally rewarding in the environment, thus uncovering symmetry. Also notice how reward shaping can convert an otherwise bland signal into one having more informational value. Since our method is dependent on estimating similarities in state action space via measuring the fraction of common observed reward trails, shaping helps distinguish these pairs by making the rewards sufficiently distinct and consequently preventing spurious similarity estimates.    
 Note that the definition of $\chi_{i,l_0}$  enables finding state action symmetries even when the actions are not invariant under the symmetry transform ie. $g_s(a)\neq a \forall s,a$ (indeed, this is the case with the Cart-Pole problem where $\forall s\in S$ $g_s(Left)=Right,g_s(Right)=Left$). Previous work in ~\cite{Girgin} is unable to do this.
Finally we present the theorem which establishes the completeness of the similarity measure $\chi_{l_0,i}$
\begin{theorem}
\label{completeness}
Let $(s,a),(s',a')$ be equivalent pairs under symmetry $\chi\langle f,g_s\rangle$ which induces the coarsest partition on $\Psi$. Assuming uniform distribution over starting states for each episode run, we have:
\begin{equation}
\lim_{|episodes|\rightarrow \infty} \chi_{i,l_0}(\langle s,a \rangle ,\langle s'a'\rangle)=1,
\end{equation}
for all symmetric pairs, $\forall l_0,i\leq |S|$.
\end{theorem}
Informally completeness asserts that any state action pair which is equivalent under the given symmetry should be identifiable using the similarity measure~\ref{similarity}.
The proof of this theorem is given in Appendix~\ref{app:theorem1-proof}. We are interested in the $\chi$ that induces the coarsest partition because it leads to the highest sample efficiency, as ideally we wish to update all the pairs in a partition in parallel for a given observation. Finally, note that our method and theorem~\ref{completeness} do not strictly require symmetries at the MDP level. The method can find more general homomorphic reductions and the associated equivalence classes (we omit this generalization for brevity and conciseness).\\

\noindent\textbf{Symmetry Inclusion Priors \& Learning}: Let $Q(s,a;\theta)$ be the function approximator being used. Having found some symmetric state action pairs $\chi_{sym}$, our next task would be to use this information while training the function approximator network. Specifically we want the network to have identical outputs for symmetric state-action pairs. This can be achieved by constraining the network to learn identical representations of the symmetric pairs in its top layers. An intuitive way of moving towards this goal would be to directly use $\chi_{sym}$ for inducing hard constraints while minimizing an appropriate loss based on one-step TD (temporal difference) targets:
\small
\begin{equation}
L_{i,TD}(\theta_i)=\mathbb{E}_{\mathbb{B}}[((r+\gamma\max_{a'} Q(s',a';\theta_{i-1}))-Q(s,a;\theta_i))^2].
\end{equation}
\normalsize
Here $\mathbb{B}$ is the set of observed $(s,a,r,s')$ tuples following a $\epsilon$-greedy behavioral policy which ensures sufficient exploration. Thus the minimization problem becomes:
\begin{align*}
\underset{\theta_i}{\min}&\;\; L_{i,TD}(\theta_i) \;\;\text{s.t.}\\
Q(s,a;\theta_i) &= Q(s',a';\theta_i) \;\forall (\langle s,a \rangle ,\langle s'a'\rangle) \in \chi_{sym}.
\end{align*}
However it becomes difficult to optimize the above problem if there are too many constraints and methods like~\cite{Cotter16} might be required. Moreover since the estimated similarity pairs are not guaranteed to be true, it may be better to solve a \emph{softer} version of the problem by introducing the symmetry constraints as an additional loss term: 
\begin{equation}
L_{i,Sym}(\theta_i)=\mathbb{E}_{\chi_{sym}}[( Q(s',a';\theta_i)-Q(s,a;\theta_i)^2].
\end{equation}
The overall loss thus becomes:
\begin{equation}
\label{tc}
\begin{aligned}
 L_{i,total}(\theta_i) = L_{i,TD}(\theta_i) + \lambda L_{i,Sym}(\theta_i),
\end{aligned}
\end{equation}
where $\lambda$ is a weighing parameter for the symmetric loss. Differentiating the total loss (Eq.~\ref{tc}) with respect to the weights, we arrive at a combination of gradients coming from the two loss objectives:
\begin{align}
\label{qlu}
\nabla_{\theta_i} L_{i,TD}(\theta_i) &=
\mathbb{E}_{\pi,s} [( r+ 
\gamma\max_{a'}Q(s',a';\theta_{i-1}) -  Q(s,a;\theta_i))\nabla_{\theta_i}Q(s,a;\theta_i)], \textrm{ and }
\end{align}
\begin{align}
\label{su}
\nabla_{\theta_i} L_{i,Sym}(\theta_i) &= \mathbb{E}_{\pi,\chi_{sym}} [( Q(s',a';\theta_i)- Q(s,a;\theta_{i-1}))\nabla_{\theta_i}Q(s',a';\theta_i)].
\end{align}
In practice we use stochastic gradient descent for loss minimization using mini batches $\mathbb{B}$. Eq.~\ref{qlu} represents the familiar Q-learning gradient for function approximation. Eq. ~\ref{su} is defined so to prevent the network from destroying the knowledge gained from current episode(Alternatively we could use same targets as in~\ref{qlu} \& increase $\lambda$ gradually, we defer for conciseness). Below we present the proposed symmetric version of the DQN algorithm.
\begin{algorithm}
\caption{Sym DQN}\label{symDQN}
\begin{algorithmic}[1]
\small{
\State Initialize memory $\mathbb{D}\gets \{\}$, $P\gets\{\{root\},\{\}\}$
\State Initialize action-value function Q with random weights($\theta$) 
\For{$episode \leq M$}
\State Initialize start state
\For{{$t=1$ to $T$}}
\State With probability $\epsilon$ select action $a_t$
\State Otherwise select $a_t= argmax_{a}Q(s_t,a,\theta)$
\State Execute action $a_t$  and observe reward $r_t$ state $s_{t+1}$
\State Store transition $(s_t; s_t; r_t; s_{t+1})$ in $\mathbb{D}$
\State Sample random minibatch $\mathbb{B}$ from $\mathbb{D}$
\State Find $\mathbb{B}_s$ the batch of symmetric pairs of $\mathbb{B}$ from $P$
\State Set targets for $\mathbb{B}$ \& $\mathbb{B}_s$
\State Perform gradient descent step as in Eq.~\ref{qlu},\ref{su}  
\EndFor
\State Update $P$ with the episode
\EndFor
}
\end{algorithmic}
\end{algorithm}

\section{Experiments}
In order to validate our approach, we compare the performance of different agents using fully connected deep feed forward neural networks with with two hidden layers for $Q$ function approximation. ReLU (rectified linear unit) non-linearity is used for activation at each hidden node.The parameters involved in symmetry learning framework were found using grid search.\\
\begin{table}
\centering
\caption{Iterations to convergence(rounded) Grid-World}
\begin{tabular}{|c|c|c|c|c|c|} \hline 
$\textbf{Setup}$   & \textbf{Naive}   & \textbf{Girgin et al.}   & \textbf{Sym}   \\ \hline
{\textbf{a}}   & {$136\pm7$}   & {$113\pm5$}    & {$\mathbf{37\pm6}$}  \\ \hline
{\textbf{b}}   & {$237\pm6$}   & {$166\pm7$}    & {$\mathbf{46\pm6}$}   \\ \hline
{\textbf{c}}   & {$174\pm7$}   & {$116\pm6$}    & {$\mathbf{31\pm5}$}   \\ \hline
{\textbf{d}}   & {$241\pm5$}   & {$187\pm6$}    & {$\mathbf{38\pm6}$}   \\ \hline
{\textbf{e}}   & {$253\pm8$}   & {$177\pm6$}    & {$\mathbf{41\pm7}$}   \\ \hline
{\textbf{f}}   & {$275\pm9$}   & {$229\pm10$}    & {$\mathbf{54\pm8}$} \\ \hline
\end{tabular}
\label{gwtable}
\end{table}
\begin{figure*}
\centering
\includegraphics[width=\textwidth]{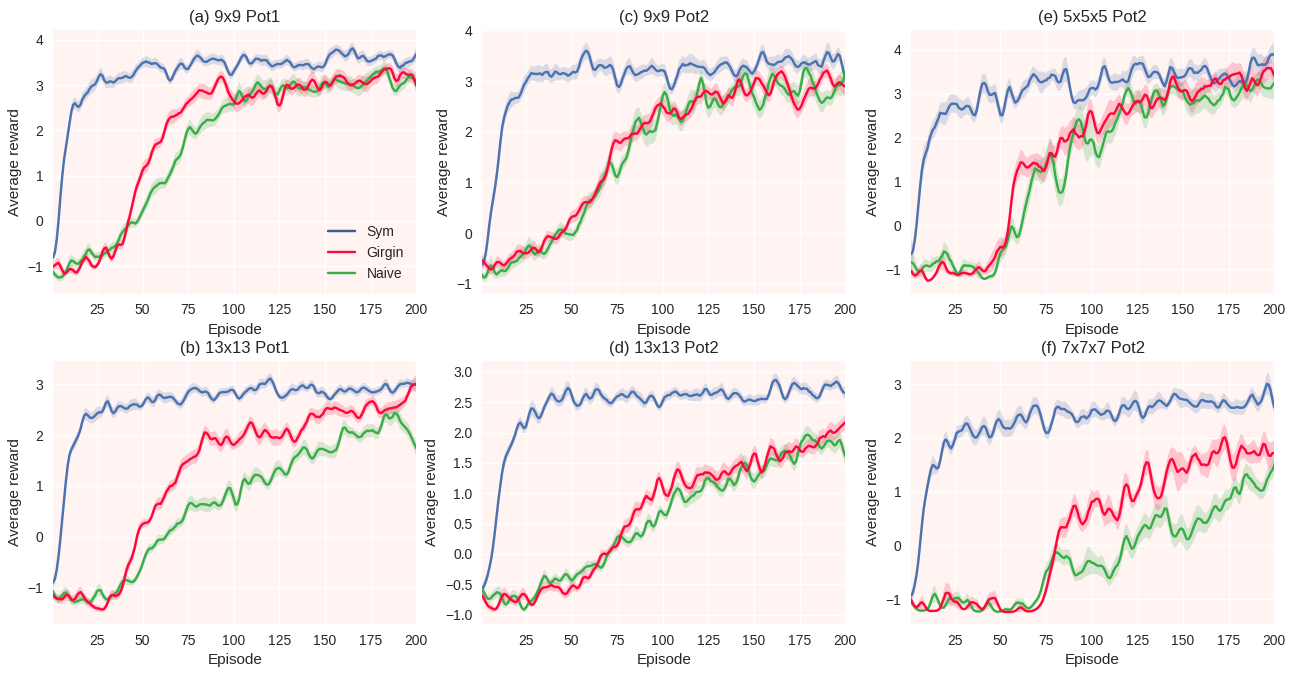}
\caption{Grid-World plots with one standard deviation errors. $\Delta=0.8,l_0=1,i=5,\lambda=0.4$. No of iterations: $50$.}
\label{2d}
\end{figure*}
\begin{table*}
\centering
\caption{Performance and statistical significance : Cart-Pole Domain}
\resizebox{.99\textwidth}{!}{
\begin{tabular}{|c|c|c|c|c|c|c|c|c|} \hline 
\multicolumn{3}{|c|}{} & \multicolumn{3}{c|}{\textbf{mean}} & \multicolumn{3}{c|}{\textbf{max}}\\ \hline
{$\mathbf{l_0}$}   & {$\mathbf{i}$}   & {{$\mathbf{\Delta}$}}    & $\textbf{DQN}$   & $\textbf{SymDQN}$   & $\textbf{P-value}$   & $\textbf{DQN}$   & $\textbf{SymDQN}$   & $\textbf{P-value}$\\ \hline
{$1$}   & {$5$}   & {$0.8$}    & $102.83\pm 69.74$  & $\mathbf{262.74\pm 64.59}$ & $<10^{-5}$ & $537.38\pm 155.89$    & {$\mathbf{796.23\pm 173.65}$} & $\num{2e-4}$\\ \hline
{$1$}   & {$5$}   & {$0.5$}    & $93.47\pm 62.16$  & $\mathbf{233.92\pm 69.48}$ & $<10^{-4}$ & $481.60\pm 163.25$    & $\mathbf{812.07\pm 181.91}$ & $<10^{-4}$\\ \hline
{$1$}   & {$4$}   & {$0.8$}    & $107.96\pm 67.17$ & $\mathbf{236.14\pm 76.12}$ & $<10^{-4}$ & $561.27\pm 156.97$    & $\mathbf{759.36\pm 168.32}$ & $\num{2.4e-3}$\\ \hline
{$2$}   & {$5$}   & {$0.8$}    & $81.57\pm 53.92$  & $\mathbf{232.65\pm 89.73}$ & $<10^{-4}$ & $514.53\pm 161.58$    & $\mathbf{860.17\pm 192.71}$ & $<10^{-4}$\\ \hline
{$2$}   & {$5$}   & {$0.5$}    & $115.24\pm 57.35$ & $\mathbf{257.02\pm 58.49}$ & $<10^{-4}$ & $467.92\pm 183.36$    & $\mathbf{671.87\pm 169	.37}$ & $\num{3.7e-3}$\\ \hline
{$2$}   & {$4$}   & {$0.8$}    & $91.33\pm 71.39$  & $\mathbf{229.48\pm 80.56}$ & $<10^{-4}$ & $503.76\pm 162.13$    & $\mathbf{807.49\pm 177.18}$ & $<10^{-4}$\\ \hline \end{tabular}
}
\label{cptable}
\end{table*}
\begin{figure*}
\centering
\includegraphics[width=\textwidth,height=5cm]{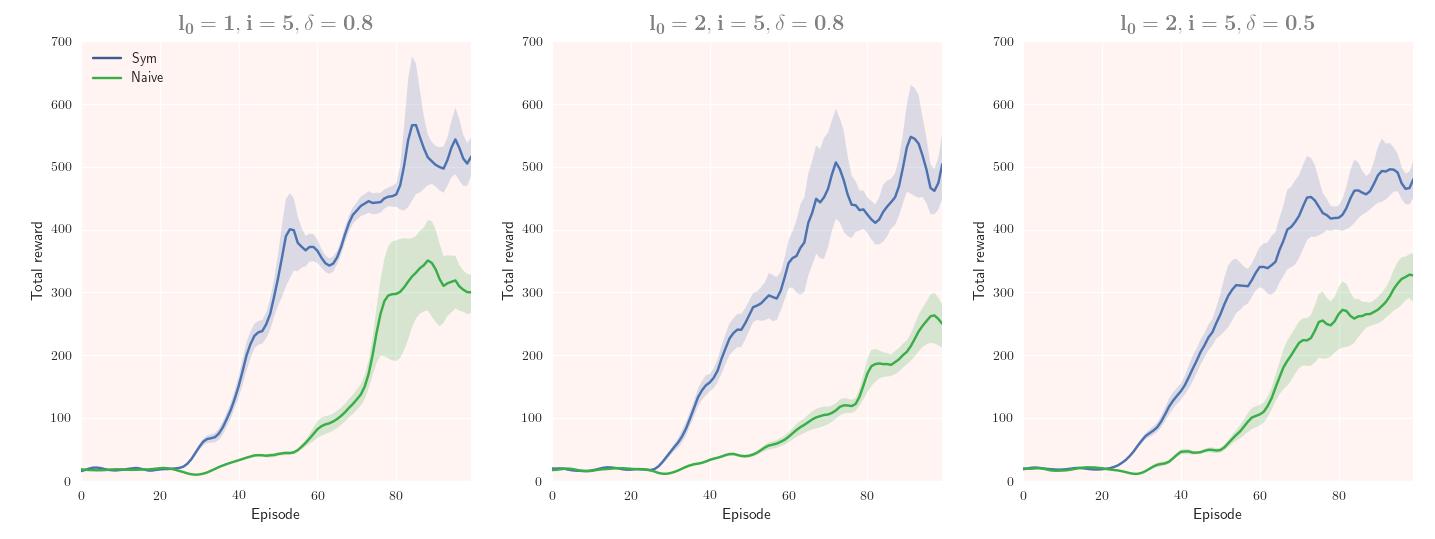}
\caption{Total reward with episodes. All experiments used $\mathbf{\lambda=1,\gamma=0.99,\epsilon}$ was decayed to 0.1 starting from 1 at rate 0.98.}
\label{cpcplot}
\end{figure*}

\noindent\textbf{Grid World}:
In the grid world domain we consider the problem of finding the shortest path to a goal in a $nXn$ square grid world. The state space is described by the coordinates $(x,y), 1\leq x,y\leq n$. The action set is $\{N,E,W,S\}$ corresponding to the directions in which the agent can move. Upon taking a desired step, we have a $90\%$ chance of landing on the expected state and a $10\%$  chance of landing on a randomly chosen adjacent state. Each episode starts at a randomly chosen state . The discount factor is set to be $\gamma = 0.9$ and exploration $\epsilon=0.1$. The goal state($x_G,y_G$) is chosen randomly at the start of each iteration. For the $2$D grid world we test two kinds of reward shaping settings, first one based on the potentials: $\Theta_1(x,y) = (|x-x_G|+|y-y_G|)$ (referred to as $Pot 1$), and second using: $\Theta_2(x,y) = (|x-x_G|+|y-y_G|)\gamma^{|x-x_G|+|y-y_G|}$ (referred to as $Pot 2$). The maximum episode length for grid sizes $9$ and $13$ are set to $480$ and $800$ respectively.The input for Grid-World net is a sparse coding of size $|S|+|A|$. Architecture used is $input\rightarrow 120\rightarrow 40\rightarrow 1$.The average reward per time step for both the potential settings are plotted in Fig.~\ref{2d}(a,b,c,d). A comparison with the previous work by Girgin et al.~\cite{Girgin} and a naive agent which uses no symmetry information is also included.  

Table~\ref{gwtable} gives the number of episodes required for the convergence to optimal policy for each setup in Figure~\ref{2d}. The agent running our algorithm (labeled $Sym$) learns the optimal policy much faster than baseline (labeled $Naive$) and previous work. We performed the two sided Welsh's t-test to check if the performance of the agents using our approach is same as that of agents using other methods. For both the scenarios("Sym vs Girgin et al."\& "Sym vs Naive") the p-values are $< 10^{-4}$ for all the setups, this shows that the difference in performance are extremely significant and Alg.~\ref{symDQN} is indeed capable of using symmetry information efficiently. Next, reward shaping potentials $Pot 2$ are more informative than $Pot 1$ as they also convey nearness to the goal state, thus agents learning under former tend to converge faster (a,b vs c,d). Finally, it is also evident that the effects of adding symmetry priors for learning become more significant as the size of the grid increases(a,c vs b,d). 
We next analyze how our method performs when the dimensionality of the grid is increased, we set up a $3D$ grid world in similar vein. Notice that this domain has $3!2^3=48$ fold symmetry. $Pot2$ type potentials are used : $\Theta(x,y,z) = (|x-x_G|+|y-y_G|+|z-z_G|)\gamma^{|x-x_G|+|y-y_G|+|z-z_G|}$. The maximum episode length for grid sizes $5$ and $7$ are set to $1000$ and $1500$ respectively. Architecture used is $input\rightarrow 300\rightarrow 120\rightarrow 1$. The average reward per time step are plotted in Fig.~\ref{2d}(e,f). It is evident form the plot and table~\ref{gwtable} that increase in dimensionality impacts performance of our method relatively mildly in comparison to other methods which fail to converge within $200$ episodes in (f). This makes intuitive sense as for a grid of size $n$ in $d$ dimensions, we have a total of $O(2dn^d)$ Q values to learn, whereas if we were to use symmetries for model reduction, we would be only be required to learn $O(\frac{2dn^d}{d!2^d})$ values.\\


\noindent\textbf{Cart-Pole}:
The agent's state here is given by $(\theta,x,\omega,v)$ denoting the angular position of pole, position of cart and their time derivatives is continuous and is typically bound by a box in $4$ dimensions. The action space is the move set $\{Left, Right\}$ for the cart. The agent gets a reward of $+1$ after every time step it manages to keep the pole balanced within the box bounds. As discussed before the state action pairs $\langle(\theta,x,\omega,v),Left\rangle$  \& $\langle(-\theta,-x,-\omega,-v),Right\rangle$ form an equivalence class under the symmetry exhibited by the domain. Notice the difficulty in finding similar state-action pairs using only the observed rewards in the conventional reward setting:  very long reward histories will be needed and the estimates will have many false positives. We define a reward shaping function for this domain as follows. To keep the symmetry finding tractable we discretize the state space into $L$ levels along each dimension using uniformly spaced intervals, This is done only for reward assignment (the agent still gets continuous input observations for the state). Thus, if the bounding box for position were $[-X_{b},X_{b}]$, then each interval is of width $w=\frac{2X_{b}}{L}$ and the discrete dimension $x_d=k$ if $\frac{(2k+1)w}{2}\geq x\geq \frac{(2k-1)w}{2}$. The shaping function is defined as : $F(\theta,x,\omega,v)=1-\frac{(\theta_d^2+x_d^2+\omega_d^2+v_d^2)}{(L-1)^2}$. Intuitively, this shaping motivates the agent to keep its coordinates near stable Cart-Pole configurations. We modify the `CartPole-v0' environment available in the OpenAI Gym platform~\cite{brockman2016openai} for our experiments. The algorithms we experimented with are the DQN~\cite{mnih2015human} agent and its proposed symmetric variant Symmetric DQN (Algorithm ~\ref{symDQN}) agent. We use a discretization level $L=9$ for the experiments, The maximum episode length is set to $1500$ the replay memory size is set $100000$ and the mini batch size of $128$ is used. Agents are run under a completely random policy for the first $25$ episodes to ensure proper initialization of the replay memory. Architecture used is $4\rightarrow 100\rightarrow 100\rightarrow 2$.  

Figure \ref{cpcplot} shows the variation of total reward obtained with the number of episodes averaged over $15$ iterations for three different parameter settings. Table~\ref{cptable} gives the mean and the maximum values of the total rewards obtained in an episode for the two algorithms, SymDQN clearly performs much better in both the metrics. Once again we perform Welsh's test to measure statistical significance of the observations for both metrics(reported in Table) and find that we can comfortably reject the hypothesis that symmetry inclusion via our method has no change on performance. 

Although our main objective was to ensure sample efficiency, we would like to point out that our method compares favorably to the baseline agents in terms of training times. For grid-world they were at-most 1.43 times slower whereas for cart-pole they were at-most 1.57 times slower.

\section{Conclusion}
In this paper we have proposed a novel framework for discovering environment symmetries and exploiting them for the paradigm of function approximation, The framework consists of two main components: The similarity detecting procedure which calculates similarity estimates between state action pairs from reward observations, and the similarity incorporating component which promotes learning of symmetric policies by imposing a symmetry cost. Our approach is scalable and requires minimal additional time and space overheads, Further we have proved the completeness of our similarity measure. We have shown the efficacy our method through extensive experimentation using deep nets on Grid-world and Cart-Pole domains. Further we have shown that the benefits of using symmetry information while learning get more profound as the dimensionality of the environment increases as there is scope for multiple symmetry occurrences. Finally we also noticed the important role reward shaping played for the method.

We believe that incorporation of discovered environment symmetries as complementary knowledge in reinforcement learning is a promising direction of research and we hope to have justifiably motivated it through this work. For future work we would like to analyze the convergence properties of our framework.   
%
\appendix

\section{Proof of Theorem~\ref{completeness}} 
\label{app:theorem1-proof}
\begin{proof}
We give the proof sketch using a coupling argument. 
Given MDP $M$, Let $\{s_t^e,a_t^e,r_t^e\}_{t=0}^{T^e}$ be the random variables observed denoting states, actions and rewards respectively while following a symmetric policy $\pi$ in an episode $e$ of length $T$, Let $\hat{N_{sa}}$ be the number of times $M$ transits though the state action pair $s,a$ and $\hat{N_{\sigma,sa}}$ denote the frequency of observing reward sequence $\sigma$ after such a transit after some number of episode runs (thus  $(\sigma,\hat{N_{\sigma,sa}})\in \Pi_{sa,T^e}$). Clearly all the estimates in $\lim_{|episodes|\rightarrow \infty}\frac{\hat{N_{\sigma,sa}}}{\hat{N_{sa}}}$ will converge in distribution to the true marginals.

Now the assumption of uniform distribution over starting state allows us to define a coupled policy execution: For every episode $e$ starting with $s_0^e,a_0^e$ we begin an episode $e'$ starting with $f(s_0^e),g_{s_0^e}(a_0^e)$. Further, for each action $a_t^e=\pi(s_t^e)$ taken in $e$ we take action $a_t^{e'}=g_{s_t^e}(a_t^e)$ in $e'$ and force the transition to $s_{t+1}^{e'}=f(s_{t+1}^e)$.Clearly both the sequences obey the transition dynamics of $M$ as the former is driven by it and the latter conforms due to given symmetry equivalence ~\ref{symtr} and thus form a coupling $\mathbb{C}(\{s_t^e,a_t^e,r_t^e\}_{t=0}^{T^e},\allowbreak\{s_t^{e'},a_t^{e'},r_t^{e'}\}_{t=0}^{T^{e'}})$. Since we had started from symmetric state action pairs and take symmetrical transition from then on by ~\ref{symrr} the reward sequence observed must be exactly identical $(r_0^e,..r_{T^e}^e)=(r_0^{e'},..r_{T^e}^{e'})$, hence all reward based observations for symmetric pairs must be identical in particular: $\hat{N_{\sigma,sa}}=\hat{N_{\sigma,s'a'}}$, Finally from the definition ~\ref{similarity} we have $\chi_{i,l_0}(\langle s,a \rangle ,\langle s'a'\rangle)=1$ in the coupled execution at all the times. Since $\mathbb{C}$ and the uncoupled case converge in distribution to same behavior in $\lim_{|episodes|\rightarrow \infty}$ the theorem is proved.    
\end{proof}

\section{Definitions}
Let $X,Y$ be sets, element $x\in X$, we then have the following constructs:
\begin{definition}
Partition: $B:=\{b_i|b_i\subseteq X\}$ is a partition of $X$ iff $ (\cup_i b_i=X) \land (b_i\cap b_j = \emptyset|i\neq j) $. We denote the block to which $x$ belongs by $[x]_B$.
\end{definition}
\begin{definition}
Coarseness: Let $B_1,B_2$ be two partitions of $X$ then we say $B_1$ is coarser that $B_2$ written $B_1\geq_c B_2$ iff $\forall x,x' \in X$, $[x]_{B_2}=[x']_{B_2} \Rightarrow [x]_{B_1}=[x']_{B_1}$.   
\end{definition}
\begin{definition}
Projection: Let $B$ be a partition of $Z\subseteq X*Y$, let $B(x)$ denote the distinct blocks of $B$ containing pairs of which x is a component, we define $B|X$ the projection of $B$ onto $X$ as the partition of $X$ which follows: $x,x'\in X \land ([x]_{B|X}=[x']_{B|X}) \iff B(x)=B(x')$.  
\end{definition}
\label{app:defs}
\bibliographystyle{myIEEEtran}
\bibliography{ijcai17}

\end{document}